\documentclass[10pt,twocolumn,letterpaper]{article}
\pdfoutput=1
\usepackage{iccv}
\usepackage{times}
\usepackage{epsfig}
\usepackage{graphicx}
\usepackage{amsmath}
\usepackage{amssymb}

\usepackage{multirow}
\usepackage{float}
\usepackage{color}
\usepackage{xcolor}
\usepackage{colortbl}
\usepackage{url}
\usepackage[american]{babel}

\usepackage[pagebackref=true,breaklinks=true,letterpaper=true,colorlinks,bookmarks=false]{hyperref}

\iccvfinalcopy 

\ificcvfinal\fi

\begin{document}
	
	\title{PAFNet: An Efficient Anchor-Free Object Detector Guidance}
	\author{Ying Xin\thanks{Both authors contributed equally to this work.}, Guanzhong Wang$^*$, Mingyuan Mao, 
		   \\Yuan Feng, Qingqing Dang, Yanjun Ma, Errui Ding, Shumin Han \\

    {\tt\small\{xinying, wangguanzhong, v\_maomingyuan, fengyuan01, dangqingqing, }\\
     {\tt\small\ mayanjun02, dingerrui, hanshumin\}@baidu.com
    }
        \\ Baidu Inc.
    }

    \maketitle
    \begin{abstract}
    Object detection is a basic but challenging task in computer vision, which plays a key role in a variety of industrial applications. 
    However, object detectors based on deep learning usually require greater storage requirements and longer inference time, which hinders its practicality seriously. Therefore, a trade-off between effectiveness and efficiency is necessary in practical scenarios. 
    Considering that without constraint of pre-defined anchors, anchor-free detectors can achieve acceptable accuracy and inference speed simultaneously. In this paper, we start from an anchor-free detector called TTFNet, modify the structure of TTFNet and introduce multiple existing tricks to realize effective server and mobile solutions respectively.
    Since all experiments in this paper are conducted based on PaddlePaddle, we call the model as PAFNet(Paddle Anchor Free Network). For server side, PAFNet can achieve a better balance between effectiveness (42.2\% mAP) and efficiency (67.15 FPS) on a single V100 GPU. For moblie side, PAFNet-lite can achieve a better accuracy of (23.9\% mAP) and 26.00 ms on Kirin 990 ARM CPU, outperforming the existing state-of-the-art anchor-free detectors by significant margins. Source code is at \url{https://github.com/PaddlePaddle/PaddleDetection}.
	\end{abstract}

	\begin{figure*}[]
	\begin{center}
 		\centerline{\includegraphics[scale=0.4]{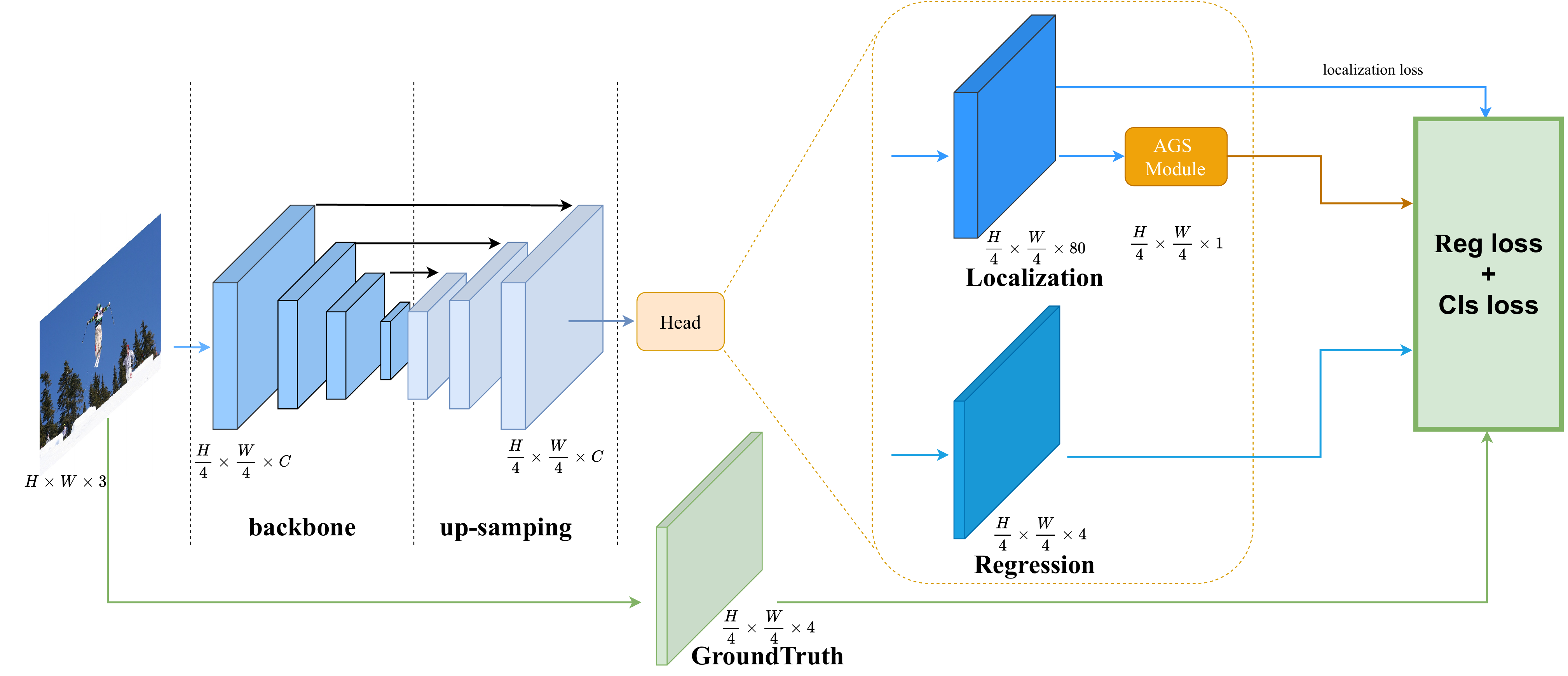}}
		\caption{The architecture of PAFNet. The overall network is composed of a backbone, an up-sampling module, an AGS module, a localization branch and a regression branch. Specifically, we choose ResNet50-vd~\cite{ResNet16} as the backbone for server side, and MobileNetV3~\cite{mobilenetv3} for mobile side. Besides, for mobile side, we replace traditional convolution layers with lite convoluation operations as shown in Fig.~\ref{Fig3}}
		\label{Fig1}
	\end{center}
    \end{figure*}
    
	\section{Introduction}
	The performance of object detectors has been dramatically improved due to the unprecedented representation capacity of convolutional neural networks (CNNs)~\cite{li2016survey, krizhevsky2017imagenet}. 
	In~\cite{Survey2019}, existing object detectors are generally based on anchors, which are categorized into two-stage~\cite{RCNN14,FastRCNN15,FasterRCNN15,cascade18} and one-stage methods~\cite{iff,YOLO16,Redmon_2019,SSD16}. 
	Specifically, two-stage methods have two main networks in their pipeline. One is designed to generate rough locations of target objects, called region proposals, the other is to fine-tune the locations and generate the corresponding category labels. By contrast, one-stage methods directly predict the locations and categories of the targets, which achieve end-to-end detection.
	One-stage methods are widely adopted in most practical scenarios because of its advantages in terms of storage requirements and inference speed. 
    It's worth noting that one-stage methods rely on a set of pre-defined anchor boxes, which has long been believed as the key for networks to converge. Nevertheless, large amounts of anchors hamper the generalization ability of detector, and increase the amount of computation and memory significantly.  
    
    Anchor-free detectors~\cite{kong2019foveabox,CenterNet2019,tian2019fcos,ttfnet2020} are proposed to address these issues by removing pre-defined anchors and regressing the locations directly, which can achieve higher efficiency. Among anchor-free detectors, the performance of TTFNet~\cite{ttfnet2020} achieves good balance between accuracy and efficiency. 
    
    In this paper, we try to explore how to appropriately introduce effective existing strategies to TTFNet~\cite{ttfnet2020} without sacrifying the efficiency, and acquire a detector for both server and mobile side that meet practical requirements of industrial applications. 
    For server side, in detail, we use ResNet50-vd as the backbone and implement SSLD, a semi-Supervised learning method to do knowledge distillation. On the detector head, we introduce a specially designed attention module (AGS), which proves to be pretty effective. For data augmentation, we finally choose CutMix~\cite{cutmix} according to the experimental results of various augmentation methods. Besides, we try 
   1x, 4x and 10x training schedule, and implement EMA strategy during training.
	For mobile side, we choose MobileNetV3-Large~\cite{mobilenetv3} as the backbone, aiming to reduce the computation cost and memory. Similar to the server side, we also implement SSLD for backbone and introduce lite structure to the head. For data augmentation, we implement CutMix~\cite{cutmix}, GridMask~\cite{grid-mask} and refer to methods from PP-YOLO~\cite{long2020pp} such as Random-Expand and Random-Crop to improve detector performance. Besides, we try 1x and 20x training schedule during training.  
	
	According to the experiments, PAFNet model for server side improves the mAP on MSCOCO 2017~\cite{Lin2014MicrosoftCC} validation set from 34.3 \% to 42.2\% with 67.15 FPS, meanwhile PAFNet-lite for mobile side achieves a better accuracy of 23.9\% mAP and 26.00 ms on Kirin 990 ARM CPU. The code and models are released in the PaddleDetection code-base (\url{https://github.com/PaddlePaddle/PaddleDetection}).
	
	\section{Related Work}
	Object detection methods based on deep learning can be roughly divided into anchor-based~\cite{YOLO9000,SSD16,cascade18,iff}, and anchor-free detectors~\cite{kong1904foveabox,CornerNet2018,CenterNet2019,tian2019fcos,ttfnet2020}. Anchor-based detectors mainly consist of two-stage and one-stage object detection methods, while anchor-free detectors consist of center-based and keypoint-based object detection methods.

    Anchor-based detectors adopt anchor boxes as pre-defined proposals for bounding box regression, which have been the mainstream in object detection field for a long time.
    However, anchor-based detectors suffer several drawbacks. First of all, the utilization of pre-defined anchors introduce additional hyper-parameters and large amounts of computation, which slow down the training and inference speed. Besides, most anchors are actually only backgrounds, leading to the class imbalance problem. Finally, the use of anchor impair its generalization ability to other datasets.
    In recent years, anchor-free methods such as FoveaBox~\cite{kong2019foveabox}, CornerNet~\cite{CornerNet2018}, CenterNet~\cite{CenterNet2019}, and FCOS~\cite{tian2019fcos} are proposed to solve aforementioned problems, which show greater potential than the SOTA anchor-based detectors.
    
    \subsection{CenterNet}
    To accelerate the inference stage, CenterNet represents objects by a single point at their bounding box center. Other properties, such as object size, dimension, 3D extent, orientation, and pose are then regressed directly from image features at the center location, and thus converting detection tasks to a standard keypoint estimation problem. The center point of objects are acquired in the format of a heatmap generated by the network, and the peaks of the heatmap are regarded as the center points of  the targets. Such strategy is a single network forward-pass and abandon the tedious post-processing procedure~\cite{nms2006,soft-nms2017,nmw2017,nmw2017-2,wbf2019}, which greatly shorten the inference time. However, CenterNet merely focuses on the object center for size regression, loses the opportunity to utilize the information near the object center, which slows down the network convergence. CenterNet needs 140-epochs training on public dataset MSCOCO~\cite{Lin2014MicrosoftCC}. In contrast, the other types of network usually requires 12 epochs.
    
    \subsection{TTFNet}
    From the aspect of practical application,  TTFNet~\cite{ttfnet2020} further improves CenterNet and achieves better balance between accuracy and efficiency. Aiming to shorten the training time, TTFNet proposes a novel approach using Gaussian kernels to encode training samples for both localization and regression. Specifically,  Gaussian probabilities are treated as the weights of the regression samples to emphasize those samples near the object center. It allows networks to make use of much more positive samples and thus converge faster with larger learning rate. Together with the light-head, single-stage, and anchor free designs, TTFNet achieves a good balance among training time, inference speed, and accuracy.
    
    Based on TTFNet, we experiment on various effective tricks to achieve better performance in terms of mAP without sacrifice on inference time both for server side and mobile side.
    
    
    

	\section{Method}
	In this section, we first introduce details of the proposed network structure for server side and mobile side, respectively. Next, we briefly describe the existing methods to implement an effective and efficient detector.
	
	\subsection{Architecture}
	\textbf{PAFNet for server side.} The overall network architecture of PAFNet is shown in Fig.~\ref{Fig1}. It is a simple and efficient network consisting of backbone, up-sampling module, AGS module, localization branch and regression branch.
    
\begin{figure}[h]
		\begin{center}
			\centerline{\includegraphics[scale=0.25]{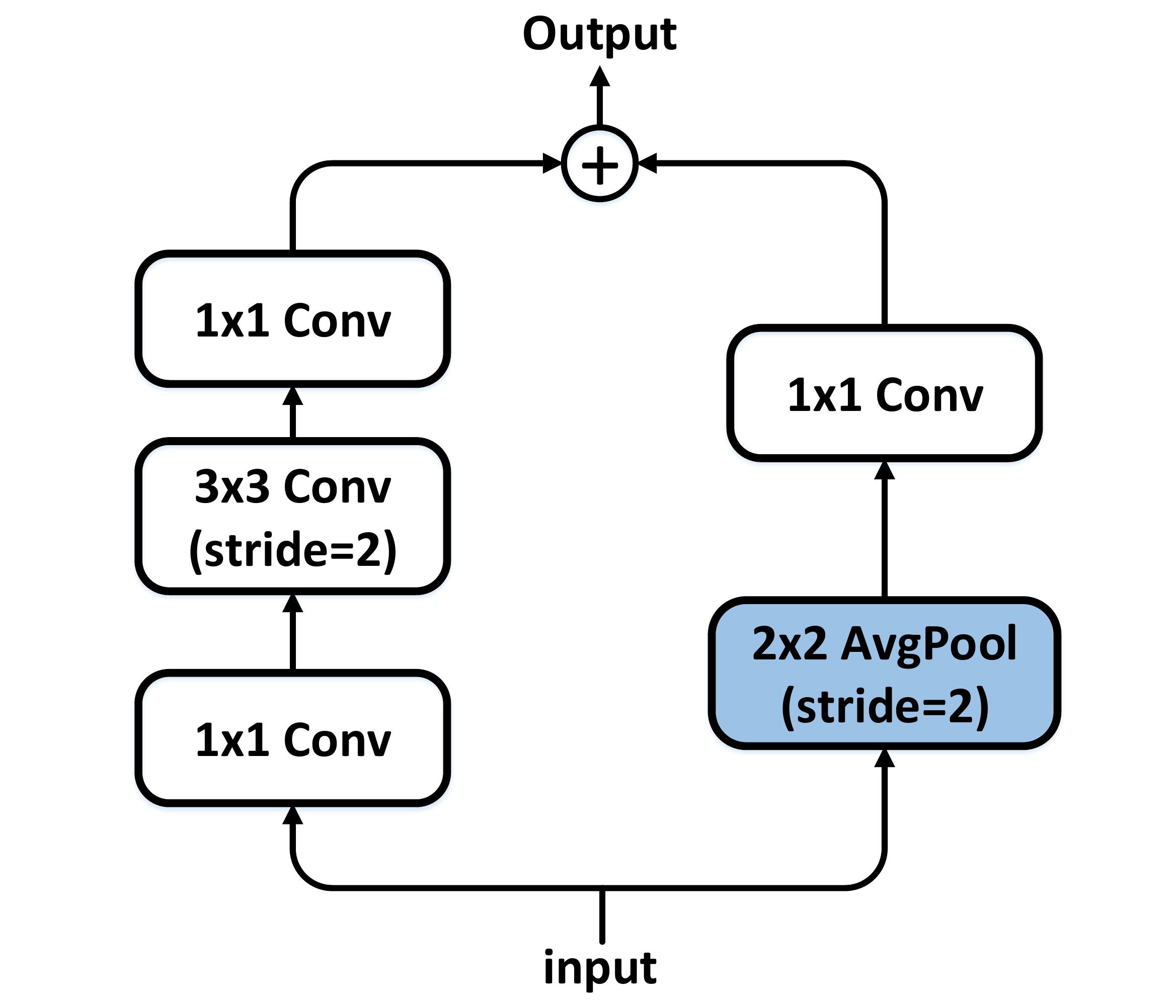}}
			\caption{The architecture of ResNet-vd module. The blue part represents the extra average pooling layer.}
			\label{Fig2}
		\end{center}
    \end{figure}

    The backbone is responsible for extracting feature over an input image with the shape ${H \times W \times 3}$. We adopt ResNet50-vd~\cite{long2020pp} as the backbone, which has better performance than ResNet50~\cite{theckedath2020detecting} on ImageNet~\cite{deng2009imagenet} dataset and maintains nearly the same inference speed. Different from ResNet50, ResNet50-vd add a $2 \times 2$ average pooling layer at each residual module (see Fig.~\ref{Fig2}.) The feature maps at layer ${i=2,3,4,5}$ of a backbone is denoted as ${F_i}$. By using a decoupling operation, the features are up-sampled to a ${1/4}$ resolution of input image size. To improve the accuracy of small objects, the shortcut connections are added between the feature ${F_i}$, ${i=2,3,4}$, and corresponding up-sampling network. After that, by adopting three convolutional layers after up-sampling module, the detection head transforms the features for localization and regression task, respectively.
    
    Localization branch generates feature map $F_L$ with the shape of ${H/4 \times W/4 \times C}$, where C presents classes of the dataset. For Ground Truth box ${G_m}$ belongs to class ${C_m}$, we map ${G_m}$ to the size of the feature map $F_l$, called ${G_m'}$. Then we use a 2D Gaussian kernel $e^{-((x-x_0)^2+(y-y_0)^2)}$ to generate a heatmap for ${G_m'}$. The peak of the Gaussian distribution is treated as a positive target, while any other pixel is treated as a negative target. We calculate a modified version Focal Loss~\cite{lin2017focal}${L_{loc}}$ for Ground Truth heatmap ${G_m'}$ and $F_l$ at the corresponding ${C_m}$ channel to supervise the training of the localization branch. 

    Regression branch gets a feature map $F_R$ with the shape of ${H/4 \times W/4 \times 4}$. We also calculate a Gaussian distribution ${G_m}$ for the Ground Truth box. We need to calculate giou ~\cite{rezatofighi2019generalized} between all the points with in the Gaussian kernel and the feature map $F_R$. Then we use a GIoU loss ${L_{reg}}$ related to response of the Gaussian distribution ${G_m}$. Besides, inspired by TTFNet, we also introduce a sample weight ${{\omega}_g}$ for each sample according to their area to balance the contribution of objects with different size.
    So the formula of the regression loss for a sample can be described as follow:
    \begin{equation}
    \label{eq1}
    l_{reg} = (1 - giou) \times {\omega}_g.
    \end{equation}
    where $l_{reg}$ belongs to ${L_{reg}}$,  ${\omega}_g$ is the sample weight.

    The total loss of the network is defined as follows:
    \begin{equation}
    \label{eq2}
    L = {\omega_{loc}} {L_{loc}} + {\omega_{reg}}{L_{reg}}.
    \end{equation}
    Where ${L_{loc}}$ is modified by Focal Loss, $L_{reg}$ is modified by GIoU loss, ${\omega_{loc}}$ is equal to ${1.0}$ and ${\omega_{reg}}$ is equal to ${5.0}$ in our work.
    
    We also introduce AGS module to emphasize on how to calculate a probability distribution map unrelated to category differences in the localization branch and use it on regression branch adaptively. In detail, the regression loss of PAFNet is a weighted sum of size regression losses of all pixels in one target area, where the weights are provided by Gaussian kernel corresponding to the target area. The AGS module is introduced to change those weights in order to keep the training process of localization branch and regression branch consistent. To obtain the significance characteristics without category differences, a max-reduce operation is utilized to find the most significant feature value along the channel dimension, and compresses the original feature map into a matrix with only a single channel. After that, the softmax is used to calculate a response on the matrix, where the larger the response value, the greater probability of the object. Finally, as the regression branch only regress for the samples in an elliptic Gaussian distribution, we mask the softmax response by the Gaussian Kernel corresponding to the current target area. 
    
    Traditionally, the size regression loss is calculated by giou. Now under the guidance of the AGS, we reweight the $giou$ to $giou'$ as shown in Eq~\ref{eq3}:
    
    \begin{equation}
    \label{eq3}
    giou' = 1 - ( (1 - \lambda) + \lambda \times S_{ags}) \times giou,
    \end{equation}
    
    where ${giou'}$ is modified by $giou$, $\lambda$ is a trade-off parameter that ranges from 0 to 1, and $S_{ags}$ denotes the probability distribution map from the AGS module. According to the experimental results, AGS module can effectively improve the precision of our network with very limited extra memory and calculations. In addition, AGS module is only used at the stage of training which will not influence the inference speed.
    
    \begin{figure}[h]
        \begin{center}
            \centerline{\includegraphics[scale=0.3]{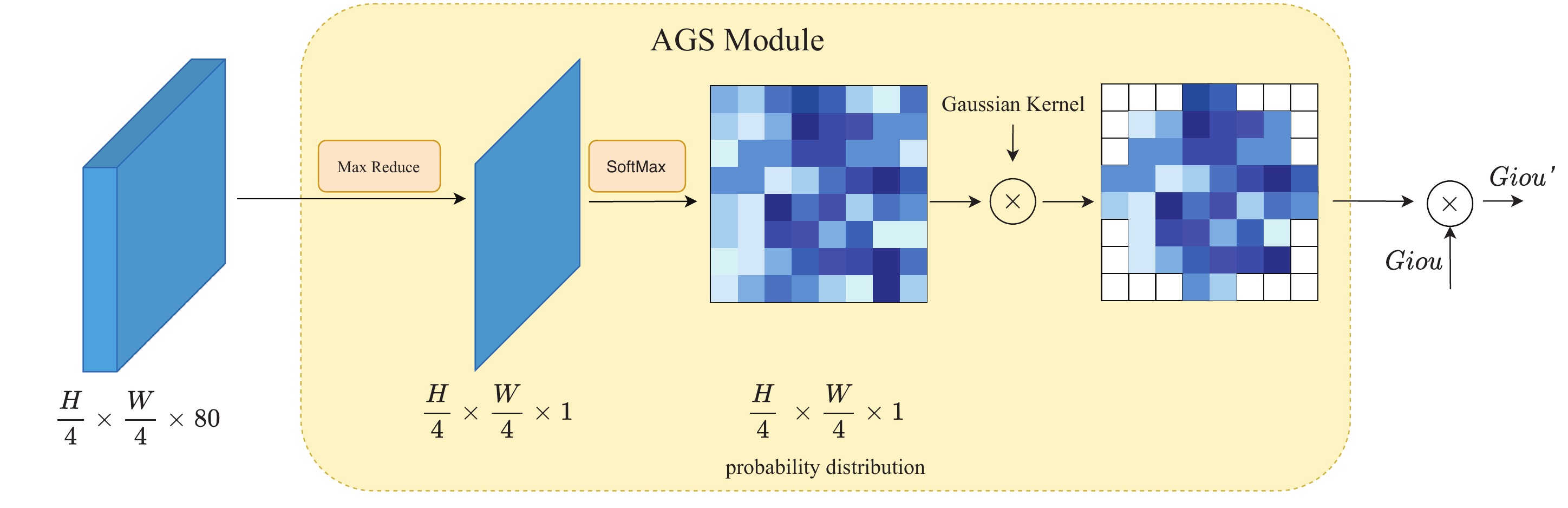}}
            \caption{The architecture of AGS Module. Output feature map of localization branch with the shape of ${H/4 \times W/4 \times C}$ is adopted as the input of AGS module. Through a series of operations including max-reduce, softmax and mask, the map is transformed to a probability distribution with the shape of ${H/4 \times W/4 \times 1}$. The output of AGS module is used to reweight the contribution of giou in regression branch.}
            \label{fig:ags} 
        \end{center}
    \end{figure}

\begin{figure}[h]
		\begin{center}
			\centerline{\includegraphics[scale=0.25]{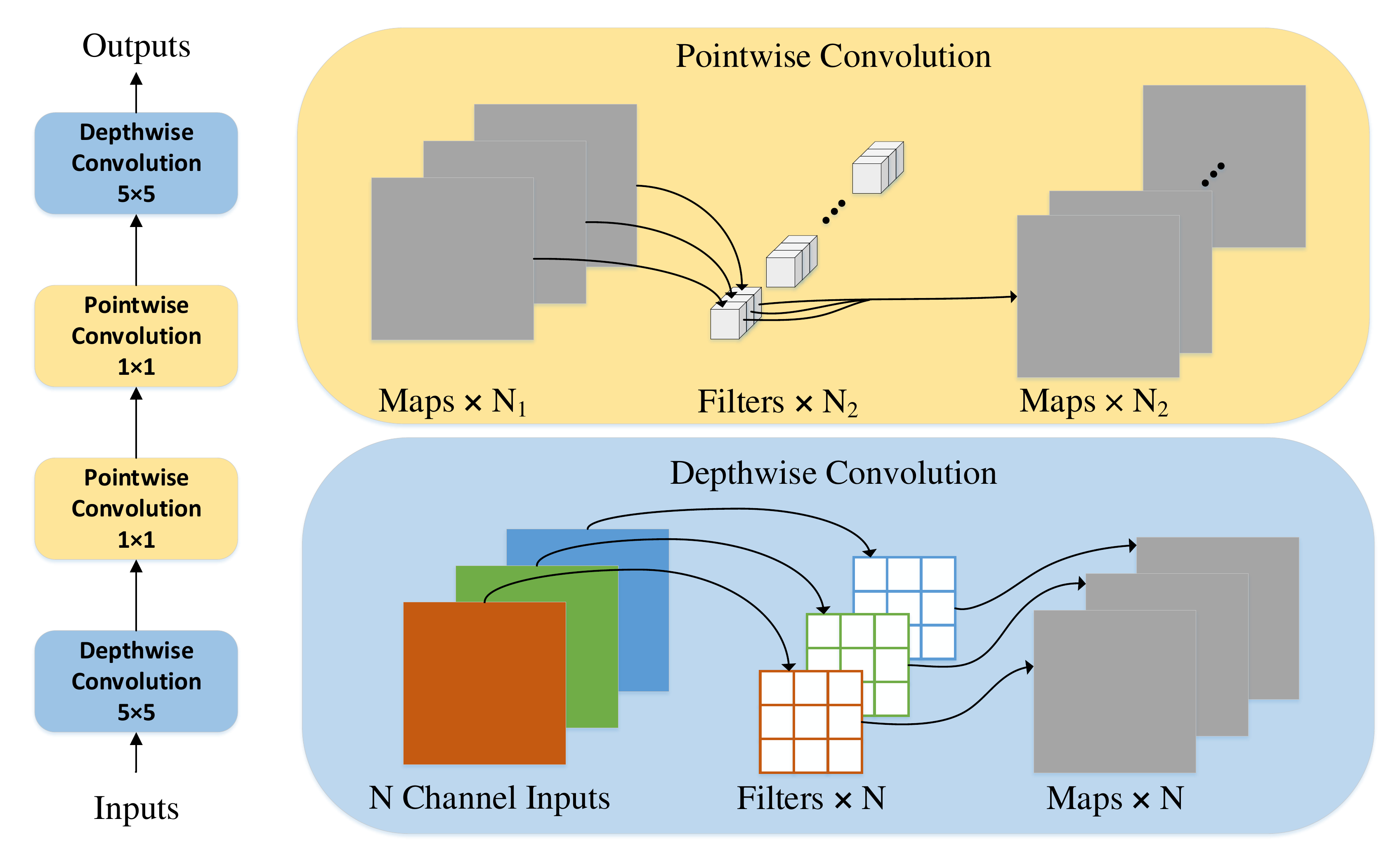}}
			\caption{The architecture of lite convolution used in up-samping network and detection head. (best viewd in color)}
			\label{Fig3}
		\end{center}
    \end{figure}

    \textbf{PAFNet-lite for mobile side.} On the mobile side, we focus more on lighter architecture and higher inference speed. Firstly, we use MobileNetV3 as backbone network, which is widely used for the models on mobile devices. Then we introduce a lite  structure, shown as Fig~\ref{Fig3} . It consists of four convolutional layers and each of them is followed by a batch-norm layer. The kernel size of these convolutional layers is 5,1,1,5, where the first and the last layers are depth-wise convolutional layer, while the second and the third are point-wise convolutional layers. The lite structure is used in up-sampling network and detection head. The number of channels on detection head is reduced from 128 to 48.
    
    \begin{table*}
    \begin{center}
    \setlength{\tabcolsep}{0.65mm}{
    \begin{tabular}{c|c|c|c|c|c|c|c|c|c|c|c|c}
    \hline
    No.  &Method                  &image size & $AP$  & $AP_{50}$ & $AP_{75}$ & $AP_S$ & $AP_M$ & $AP_L$ &Params &GFLOPS  &Infer Time &FPS           \\ \hline  \hline
    A    &TTFNet-DarkNet          &512        & 32.9  & 50.4      & 35.2      & 16.8   & 36.5   & 44.6   & 43.00M    & 134.97 & 9.79ms          & 102.14       \\ \hline
    B    &TTFNet+ResNet50vd+AGS*  &512        & 34.3  & 52.2      & 36.6      & 17.1   & 37.9   & 46.8   & 27.74M    & 104.44 & 11.72ms         & 85.42        \\
    C    &B + Better Pretrain     &512        & 36.1  & 54.7      & 38.5      & 18.8   & 39.8   & 49.6   & 27.74M    & 104.44 & 11.72ms         & 85.42        \\ 
    D    &C + 4x Scheduler        &512        & 36.5  & 54.2      & 38.7      & 19.2   & 39.8   & 50.1   & 27.74M    & 104.44 & 11.72ms         & 85.42        \\ 
    E    &D + CutMix + EMA        &512        & 36.9  & 54.6      & 39.5      & 19.3   & 40.2   & 49.8   & 27.74M   & 104.44 & 11.72 ms        & 85.32        \\ 
    F    &E + DCN                 &512        & 39.8  & 57.2      & 41.5      & 20.6   & 42.9   & 54.5   & 17.18M    & 75.21  & 14.89ms         & 67.15        \\ 
    \bf{G}   &\bf{F + 10x Scheduler}   &\bf{512}        & \bf{42.2}      &  \bf{59.8}  & \bf{45.3}  & \bf{22.8}         & \bf{45.8}     & \bf{59.2}   & \bf{17.18M}    & \bf{75.21}  & \bf{14.89ms}         & \bf{67.15}        \\ \hline
    \end{tabular}
    }
    \end{center}
    \caption{The ablation study of tricks on the MS COCO test-dev for server side.}
    \label{table1}
    \end{table*}

    \begin{table*}
    \begin{center}
    \setlength{\tabcolsep}{0.65mm}{
    \begin{tabular}{c|c|c|c|c|c|c|c|c|c}
    \hline
    No.  &Method                      &image size & $AP$ & $AP_{50}$ & $AP_{75}$ & $AP_S$ & $AP_M$ & $AP_L$ & Latency on ARM 4xCore(ms)  \\ \hline  \hline
    A    &TTFNet-MobileNet-Large      &320        & 16.3 & 30.5      & 15.9      & 4.0    & 15.4   & 28.3   & 26.00                            \\
    B    &A + Better Pretrain         &320        & 18.0 & 33.2      & 17.7      & 4.5    & 17.7   & 29.7   & 26.00                            \\
    C    &B + 20x Scheduler           &320        & 19.8 & 35.7      & 20.1      & 5.6    & 20.8   & 32.3   & 26.00                             \\ 
    D    &C + CutMix                  &320        & 20.0 & 35.9      & 20.0      & 5.6    & 20.8   & 33.3   & 26.00                              \\ 
    \bf{E}    &\bf{D + GridMask}      &\bf{320}   & \bf{23.9} & \bf{40.2}      & \bf{24.6}  & \bf{7.1}   & \bf{23.3}   & \bf{40.8}  & \bf{26.00}                               \\ \hline
    \end{tabular}
    }
    \end{center}
    \caption{The ablation study of tricks on the MS COCO test-dev for mobile side on Kirin 990 ARM CPU .}
    \label{table2}
    \end{table*}

	\subsection{Selection of Trick}
    In this section, we focus on how to combine the existing tricks~\cite{lawrance1977exponential,yun2019cutmix} for server side and mobile side to implement an effective and efficient detector.
    Considering many tricks are proposed for classification and thus can-not be applied directly, we do some modification to appropriately integrate them into the detection network.
    
    {\bf Better Pretrain Model} Using a better pretrain model with higher classification accuracy can help get a better detector. In our work, we use the distilled ResNet50-vd model, denoted as ResNet50-vd-ssld, as the pretrain model for server side, the classification accuracy of which is 83.0\%. It is obvious that the pretrain model will not affect the efficiency of the detectors. Similarly, we use distilled MobileNetV3 model as the pretrain model for mobile side. 

    {\bf EMA} Exponential moving average (EMA) is a type of moving average (MA) that gives more weight and importance to the most recent data points. It considers more information and thus becomes robust to noise to a certain extent. Exponential moving average performs well in detection tasks. Specifically, each parameter W is optimized as:
    \begin{equation}
    \label{ema}
    W_{EMA} = \lambda W_{EMA} + (1 - \lambda) W,
    \end{equation}
    
    {\bf CutMix} CutMix~\cite{yun2019cutmix} is an improved version of MixUp~\cite{zhang2017mixup} for classification. MixUp is an operation that can mix two random samples proportionally, the category of the sample is distributed proportionally as well. The formula is shown as follows:
    \begin{equation}
    \label{eq6}
    \begin{split}\small
    &x_n = {\lambda}x_i + (1 - \lambda) x_j,\\
    &y_n = {\lambda}y_i + (1 - \lambda) y_j.
    \end{split}
    \end{equation}
    Where $(x_n, y_n)$ is the new data generated by interpolation, and $(x_i, y_i)$ and $(x_j, y_j)$ are data randomly selected from the training set. The value of $\lambda$ satisfies beta distribution~\cite{gupta2004handbook}, and the value range is between 0 and 1. By contrast, according to CutMix, patches are cut and pasted among training images where the ground truth labels are also mixed proportionally to the area of the patches. By making efficient use of training pixels and retaining the regularization effect of regional dropout, CutMix usually helps achieve better performance in classification and detection. The formula is shown as follows:
    \begin{equation}
    \label{eq7}
    \begin{split}\small
    &x_n = M {\odot} x_i + (1 - M) {\odot} x_j,\\
    &y_n = {\lambda}y_i + (1 - \lambda)y_j.
    \end{split}
    \end{equation}
    Where ${M}$ is a binary mask to place filled areas, ${\odot}$ is element-wise multiplication, ${\lambda}$ belongs to the beta distribution like MixUp.

    {\bf GridMask} GridMask~\cite{chen2020gridmask} uses information deletion to enhance data. This method is implemented by randomly discarding a region on the image, which is equivalent to adding a regular term to the network to avoid network overfitting.

    {\bf DCN} Since the geometric structure of the module used to construct the convolutional neural network (CNN) is fixed, its geometric transformation modeling ability is essentially limited. Two new modules are introduced to improve the modeling capability of the convolutional neural network (CNN), including deformable convolution and deformable ROI pooling~\cite{dai2017deformable}. Both of them are based on the idea of further displacement adjustment of spatial sampled position information in the module, which can be learned from the object task and does not require additional monitoring signals.
    
    \begin{table*}
    \begin{center}
    \setlength{\tabcolsep}{0.7mm}{
    \begin{tabular}{c|c|c|c|c|c|c|c|c|c}
    \hline
    Method                               & Backbone      &image size & $AP$   & $AP_{50}$ & $AP_{75}$ & $AP_S$ & $AP_M$ & $AP_L$  &V100 FPS without TRT  \\ \hline  \hline
    CornerNet-lite~\cite{law2018cornernet} & hourglass104~\cite{li2018contextual}  &512 & 34.4   & - & - & 14.8 & 36.9  & 35.50       & 35.50               \\\hline
    EfficentDet-D0~\cite{thanh2020polyp} & Efficient-B0  &512        & 33.8   & 52.2      & 35.8      & 12.0  & 38.3    & 51.2    & 98 +                   \\
    EfficentDet-D1~\cite{thanh2020polyp} & Efficient-B1  &640        & 39.6   & 58.6      & 42.3      & 17.9  & 44.3    & 56.0    & 74.1 +                 \\
    EfficentDet-D2~\cite{thanh2020polyp} & Efficient-B2  &768        & 43.0   & 62.3      & 46.2      & 22.5  & 47.0    & 58.4    & 56.5 +                 \\
    EfficentDet-D3~\cite{thanh2020polyp} & Efficient-B3  &896        & 45.8   & 65.0      & 49.3      & 26.6  & 49.4    & 59.8    & 34.5 +              \\ \hline
    RetinaNet~\cite{lin2017focal}        & ResNet50      &640        & 37.0   & -         & -         & -      & -      & -       & 37                   \\
    RetinaNet~\cite{lin2017focal}        & ResNet101     &640        & 37.9   & -         & -         & -      & -      & -       & 29.4                  \\
    RetinaNet~\cite{lin2017focal}        & ResNet50      &1024       & 40.1   & -         & -         & -      & -      & -       & 19.6                  \\
    RetinaNet~\cite{lin2017focal}        & ResNet101     &1024       & 41.1   & -         & -         & -      & -      & -       & 15.4                  \\ \hline
    FCOS-imprv~\cite{tian2019fcos}       & ResNet50      &800       & 38.7    & -         & -         & -      & -      & -       & 22                  \\ 
    FCOS-imprv+DCN~\cite{tian2019fcos}   & ResNet50      &800       & 44.4    & -         & -         & -      & -      & -       & 13.36                \\ \hline
    CenterNet(Objects)~\cite{duan2019centernet}  &DLA-34~\cite{yu2018deep} & 512       & 37.4   & 55.1 & 40.8  & -     & -       & -      & 55.9         \\ 
    CenterNet(Objects)~\cite{duan2019centernet}  &ResNet18              & 512          & 28.1   & 44.9 & 29.6  & -     & -       & -      & 153.75       \\ \hline
    TTFNet~\cite{liu2020training}       &Darknet      & 512   & 32.9    & 50.4        & 35.2       & 16.8     & 36.5       & 44.6      & 102.14   \\ \hline
    PAFNet                           &ResNet50-vd   &512      & 42.2 & 59.8     & 45.3      &  22.8 & 45.8    & 59.2    & 67.15                 \\ \hline
    \end{tabular}
    }
    \end{center}
    \caption{PAFNet vs. other state-of-the-art anchor-based or anchor-free detectors in single-model and single-scale results for server side.}
    \label{table3}
    \end{table*}

	\section{Experiment}
    Our experiments are carried out on MS COCO dataset~\cite{lin2014microsoft} included 80 categories. Following the previous practice~\cite{girshick2015fast, kong1904foveabox}, we use the \textit{trainval35k} split containing about 115K images for training and report our main results on the \textit{test-dev} split containing about 20K images by uploading our results to the codalab.
    
    \subsection{Implementation Details}
    {\bf Training Details} For server side, we use ResNet50-vd~\cite{yu20201st} as the backbone network for our ablation study, which is initialized with the weights distilled on ImageNet~\cite{deng2009imagenet} dataset and performs better than ResNet50~\cite{akiba2017extremely}. To speed up training and inference, we resize images to  ${512 \times 512}$, which is slightly different from the common setting. In the multi-task loss function, we set the weight of localization loss as 1.0 and the regression loss as 5.0 to balance the training of two branches. What's more, the network is trained with synchronized SGD for 15K iterations with initial learning rate as 0.015 and a minibatch of 12 images distributed on 8 GPUs. The learning rate is divided by 10 at iteration 11.25K and 13.75K, respectively. Weight decay is set as 0.0004, and momentum is set as 0.9.

    For mobile side, we adopt MobileNetV3 as the backbone, which is distilled on ImageNet dataset as well. The size of input image is ${320 \times 320}$ and the other settings are similar to the server side.
    
    {\bf Inference Details} 
    At the stage of inference, we resize the input image in the same way as the training stage, and then forward it through the whole network to obtain the predicted bounding boxes with a predicted class. Since post-processing is the same as TTFNet~\cite{liu2020training}, we adopt the post-processing hyper-parameters of TTFNet directly. We hypothesize that the performance of our detector may be further improved if we carefully adjust the hyper-parameters. Detection results are evaluated with the standard COCO metric, which average mAP of IoUs across 0.5 to 0.95. Besides, the infer time of server side model is tested on a single V100 GPU with batch size = 1, and the latency of mobile side is tested on 4x Core Kirin 990 ARM CPU.

    \subsection{Ablation Study}
    In this section,  we gradually join tricks both in server side and mobile side. It's worth noting that the tricks are not completely independent, some of which work well when applied alone, but not effective when combined together. Therefore, we present how to improve the performance of the object detector step by step in the order of our exploration and discovering the effectiveness of tricks. Results are shown in Table~\ref{table1} and Table~\ref{table2}.

    \subsubsection{Ablation Study for Server Side}
    {\bf{A $\rightarrow$ B}} First of all, we follow the TTFNet with DarkNet53 as the backbone and get a basic version(A), the mAP of the model reached 32.9\%. Because the ResNet series is  widely used, we replace the original backbone Darknet53 with ResNet50-vd and apply AGS module on head. In this way, we get modified TTFNet(B) with the mAP of 34.3\%, which is higher than the original model (No.A in Tabel~\ref{table1}), while its parameters and FLOPs are much smaller than the original TTFNet model.
    
    {\bf{B $\rightarrow$ C}} Replacing the pre-trained model is a very common approach to improve model accuracy. Therefore, we use the distilled ResNet50-vd-ssld model for backbone initialization. For fair comparison, this distillation model still uses ImageNet for pre-training. The mAP of model C can be further improved by 1.8\% than the model B. In fact, using other detection datasets for pre-training can also greatly improve the performance of the model, but this is beyond the scope of this paper.
    
    {\bf{C $\rightarrow$ D}} We try to optimize the training strategy. Considering that the number of training rounds of 1x scheduler could not make the model fully convergent, we train the model with 4x scheduler and get the mAP of 36.5\%.
    
    {\bf{D $\rightarrow$ E}} We try several data augmentation methods such as Mixup, Cutout~\cite{devries2017improved} and CutMix, and the experiment results shows that CutMix get the best results. Besides, When training a model, it is often beneficial to maintain moving averages of the trained parameters. The Exponential Moving Average (EMA) compute the moving averages of trained parameters using exponential decay, for each parameter W, we maintain an shadow parameter as the Equ.~\ref{ema}. The data augmentation Cutmix and EMA policy help improve the mAP by 0.4\%.

    {\bf{E $\rightarrow$ F}} Considering that the number of parameters and FLOPs of ResNet50-vd are much smaller than those of Darknet53, we replace the 3×3 convolutional layer at the last stage of ResNet50-vd with deformable convolution layer and predict head branch. In this way, we get the model(No. F in Table~\ref{table1}) with a mAP of 39.8\%. 
    
    {\bf{F $\rightarrow$ G}} Following model $D$, we implement 10x learning scheduler and achieve the highest mAP of 42.2\%.
    
    \subsubsection{Ablation Study for Mobile Side}
    Similar to the server-side model, we also build a basic version of  network(A) with backbone MobilenetV3-large, the mAP of which is 16.3\%. Then we replace the backbone by a distilled MobilenetV3 model (B) and get a mAP of 18.0\%. In order to make the model converge enough, 20x is trained on the model and the mAP is improved by 1.8\%. The CutMix, GridMask and other data augmentation modules are also used on the mobile side and get a mAP of 23.9\% and the latency on Kirin 990 ARM CPU of 26.00 ms. 

    \subsection{Comparison with Other State-of-The-Art Detectors}
    We compare our PAFNet with other state-of-the-art anchor-free object detectors on the \textit{test-dev} split of MS COCO benchmark in Table~\ref{table3}. The FPS results of PAFNet and other methods are all tested on V100 with batch size = 1, results marked by '+' are updated results from the corresponding official code-base.
    
    As shown in Table~\ref{table3}, our PAFNet with the backbone ResNet50-vd achieves 42.2\% in AP and 67.15 FPS in speed. Compared with other state-of-the art anchor-free methods, both key-point methods and center-based methods, such as FCOS and EfficentNet, our PAFNet has certain advantages in speed and accuracy.
    
  
    
	\section{Conclusion}
	In this paper, we introduce a new implementation of anchor-free detector both for server side and mobile side. 
	For server side, PAFNet outperforms other state-of-the-art anchor-free detectors in both accuracy and speed. 
	Besides, we have optimized the framework of PAFNet-lite for mobile side. We conduct lots of experiments to figure out what tricks work on PAFNet for server and mobile side, respectively. Finally, we develop an useful anchor-free detector with good balance between accuracy and efficiency. We hope this paper can provide developers useful experimental results and detection models, and thus help them achieve better performance in practical applications.

     \section{Acknowledgements}
    This work was supported by the National Key Research and Development Project of China (2020AAA0103500).

	\bibliographystyle{ieee}
	\bibliography{paper}
\end{document}